\documentclass[letterpaper]{article} 
\usepackage{aaai25}
\usepackage{times}  
\usepackage{helvet}  
\usepackage{courier}  
\usepackage[hyphens]{url}  
\usepackage{graphicx} 
\usepackage{natbib}  
\usepackage{caption} 
\usepackage{algorithm}
\usepackage{algorithmic}
\usepackage{amsmath}
\usepackage{booktabs}
\usepackage{multirow}
\usepackage{amssymb}
\usepackage{tabularx}
\usepackage{anyfontsize}
\usepackage{subcaption}

\usepackage{newfloat}
\usepackage{listings}
\DeclareCaptionStyle{ruled}{labelfont=normalfont,labelsep=colon,strut=off} 
\floatstyle{ruled}
\newfloat{listing}{tb}{lst}{}
\floatname{listing}{Listing}
\title{Adversarial Attacked Teacher for Unsupervised Domain Adaptive Object Detection}
\author {
    Kaiwen Wang\textsuperscript{\rm 1},
    Yinzhe Shen\textsuperscript{\rm 1},
    Martin Lauer\textsuperscript{\rm 1}
}
\affiliations {
    \textsuperscript{\rm 1}Karlsruhe Institute of Technology \\
    kaiwen.wang@kit.edu, yinzhe.shen@kit.edu, martin.lauer@kit.edu
}

\begin{document}

\maketitle

\begin{abstract}
Object detectors encounter challenges in handling domain shifts. Cutting-edge domain adaptive object detection methods use the teacher-student framework and domain adversarial learning to generate domain-invariant pseudo-labels for self-training. However, the pseudo-labels generated by the teacher model tend to be biased towards the majority class and often mistakenly include overconfident false positives and underconfident false negatives. We reveal that pseudo-labels vulnerable to adversarial attacks are more likely to be low-quality. To address this, we propose a simple yet effective framework named \textit{\textbf{A}dversarial \textbf{A}ttacked \textbf{T}eacher} (\textbf{AAT}) to improve the quality of pseudo-labels. Specifically, we apply adversarial attacks to the teacher model, prompting it to generate adversarial pseudo-labels to correct bias, suppress overconfidence, and encourage underconfident proposals. An adaptive pseudo-label regularization is introduced to emphasize the influence of pseudo-labels with high certainty and reduce the negative impacts of uncertain predictions. Moreover, robust minority objects verified by pseudo-label regularization are oversampled to minimize dataset imbalance without introducing false positives. Extensive experiments conducted on various datasets demonstrate that AAT achieves superior performance, reaching 52.6 mAP on Clipart1k, surpassing the previous state-of-the-art by 6.7\%.
\end{abstract}

%

\section{Introduction}
As computer vision techniques progress, there has been a notable breakthrough in the performance of object detection \cite{fasterrcnn,yolo,cascadercnn,fcos}. However, a noticeable decline in performance occurs when these algorithms are exposed to domain shifts. To overcome this performance drop, supervised methods require extensive annotations, which is expensive and impractical. Domain Adaptive Object Detection (DAOD) \cite{dafasterrcnn,sw,umt,pt,at,cmt,cat} has thus been proposed to address this issue, where a pre-trained object detector is adapted from a labeled source domain to an unlabeled target domain. As a semi-supervised learning scheme, DAOD eliminates the need for annotating training data in the target domain, making it more practical for real-world applications. 

\begin{figure}[t]
\centering
\includegraphics[width=0.9\columnwidth]{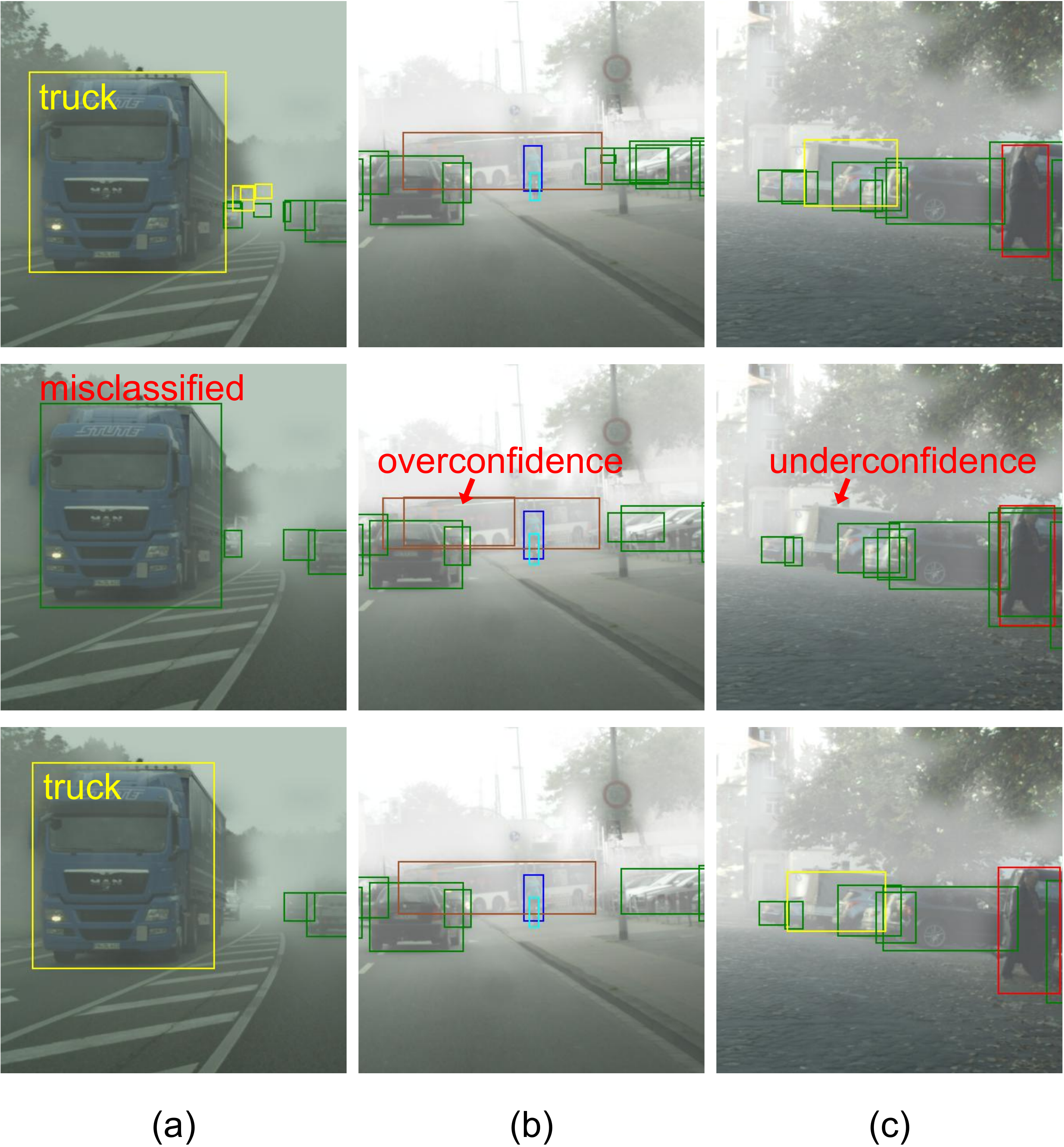} 
\caption{\textit{Ground truth} (top), \textit{vanilla pseudo-labels} (middle), and \textit{adversarial pseudo-labels} (bottom) generated on Foggy Cityscapes. Different colors represent different categories. Vanilla pseudo-labeling tends to exhibit the following issues: (a) bias towards major classes, (b) overconfidence about parts of objects, and (c) underconfidence.}
\label{fig1}
\end{figure}

The most straightforward category of DAOD involves aligning two domains through image-level or instance-level adversarial learning \cite{dafasterrcnn,sw}, aiming to acquire a domain-invariant representation. While this approach has proven effective, relying solely on adversarial learning for complex recognition tasks such as object detection still results in a significant performance gap compared to the Oracle model (fully supervised on the target domain). In recent years, the self-training paradigm \cite{umt,pt,at,cmt} has shown more promising results in mitigating domain shifts by using teacher-student mutual learning \cite{mt}. Specifically, the student model is trained to make consistent predictions as the teacher model, which is typically an exponential moving average (EMA) of the student models. However, these methods overlook that vanilla pseudo-labeling tends to be biased toward more frequently observed categories. 

Figure~\ref{fig1} illustrates several examples of low-quality pseudo-labels arising from these issues. In Figure~\ref{fig1} (a), a truck is misclassified as a more common category (car). In Figure~\ref{fig1} (b), excessive confidence is given to a part of a bus. Figure~\ref{fig1} (c) shows a false negative case, where a clearly visible truck is not included as a pseudo-label. Mutual training with such low-quality pseudo-labels leads to suboptimal performance. To address classification imbalance, class-specific thresholding has been proposed, which filters pseudo-labels by applying a lower threshold for minority classes \cite{threshold1,threshold2,threshold3}. While these approaches alleviate the imbalance of pseudo-labels to some extent and may reduce false negatives, they also tend to introduce more overconfident pseudo-labels. Additionally, Class-Relation Augmentation has been introduced to enhance the representation of minority classes by blending them with highly similar majority classes at the instance level \cite{cat}. However, this method still struggles with the presence of incorrect pseudo-labels.

In this paper, we introduce the Adversarial Attacked Teacher (AAT), a framework designed to regularize bias and address issues of false positives and false negatives in pseudo-labels for Domain Adaptive Object Detection (DAOD). AAT applies adversarial attacks on the teacher model and generates adversarial pseudo-labels on adversarial examples. By comparing vanilla and adversarial pseudo-labels, we estimate the uncertainty associated with these labels. Robust pseudo-labels that match are retained, while overconfident pseudo-labels are discarded, and underconfident predictions are incorporated. In addition to the standard consistency loss between vanilla pseudo-labels and student model predictions, the student model is trained to match adversarial pseudo-labels as a regularization term. This adaptive pseudo-label regularization has several benefits. Firstly, reliable pseudo-labels are emphasized in both the normal consistency loss and the regularization term, allowing high-quality transfer learning from confident pseudo-labels to others. Secondly, pseudo-labels biased toward dominant classes are corrected in the adversarial pseudo-labels. Together with pseudo-label regularization, this approach implicitly learns soft labels for uncertain proposals. Thirdly, overconfident pseudo-labels are removed in the adversarial pseudo-labels while underconfident pseudo-labels are added to ensure a balanced and accurate set of training labels. Additionally, robust target-domain minority objects that are resilient against adversarial attacks are cropped and stored in a Crop Bank. Those crops are oversampled to alleviate category imbalance in the dataset. This strategy enables the teacher model to learn a more balanced representation and increase the pseudo-label robustness in the further training process. Equipped with these perspectives, we summarize the following contributions.
\begin{itemize}
    \item We demonstrate several cases where low-quality pseudo-labels can negatively affect self-training, highlighting their particular vulnerability to adversarial attacks.
    
    \item We introduce Adversarial Attacked Teacher (AAT), a novel teacher-student framework that uses adversarial attack-based pseudo-label regularization and robust minority oversampling to improve pseudo-label quality. This approach emphasizes high-certainty pseudo-labels and mitigates the negative effects of uncertain predictions.
    
    \item We conduct extensive experiments to verify the effectiveness of our framework, which significantly outperforms all existing SOTA by a large margin. Our approach achieves 53.0 mAP on Foggy Cityscapes and 52.6 mAP on Clipart1k.
\end{itemize} 
\begin{figure*}[t]
\centering
\includegraphics[width=\textwidth]{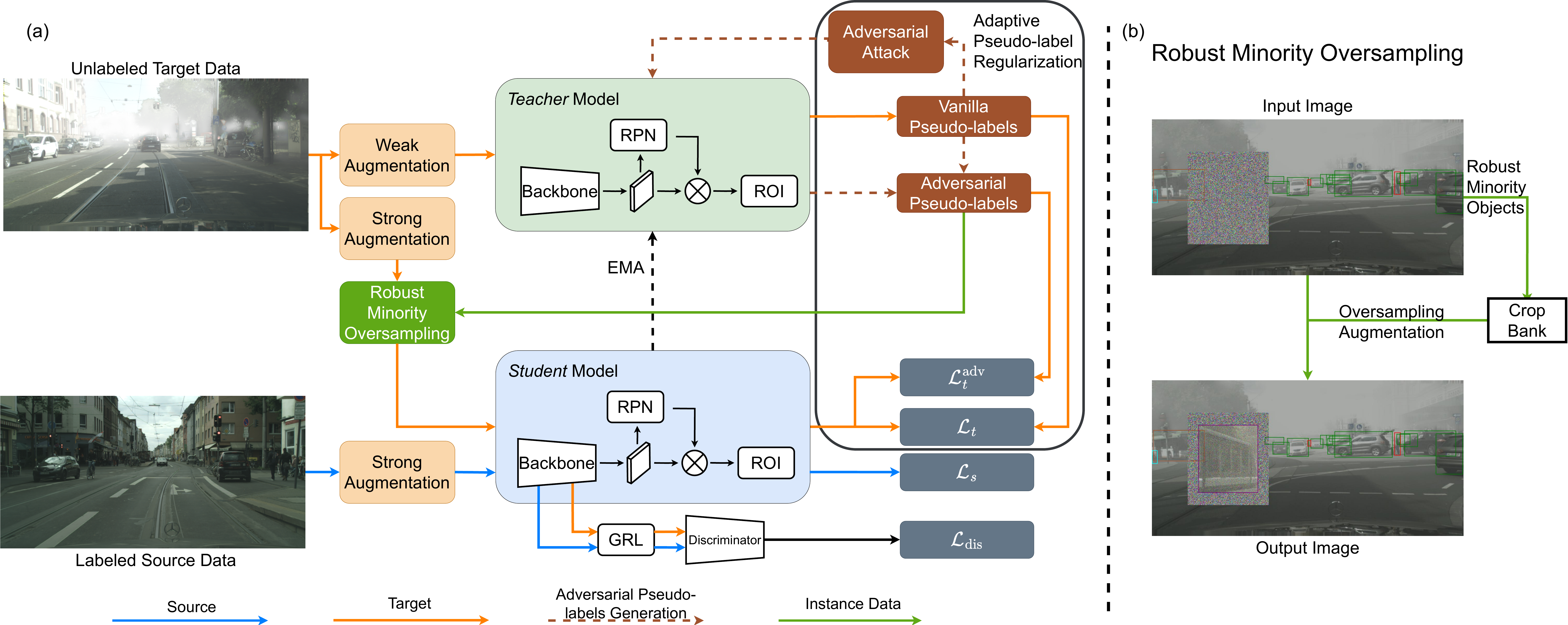} 
\caption{(a) Overview of the proposed \textit{Adversarial Attacked Teacher} framework. Our model incorporates a Mean Teacher framework and two key components: Adaptive Pseudo-label Regularization (APR) and Robust Minority Oversampling (RMO). APR involves conducting adversarial attacks on the teacher model to generate adversarial pseudo-labels, which serve as complementary labels by retaining reliable vanilla pseudo-labels and modifying uncertain ones. The student model is trained to fit both types of pseudo-labels. RMO involves oversampling high-certainty objects from minority categories to address class imbalance. (b) Robust Minority Oversampling  (RMO) demonstrated on a train object.}
\label{fig2}
\end{figure*}
\section{Related Works}
\subsection{Domain Adaptive Object Detection}
Recently, considerable work has employed domain adaptation to achieve better object detection under domain shift. These domain adaptive object detection approaches can be mainly divided into three categories: feature alignment, domain translation, and self-training. Feature alignment methods \cite{dafasterrcnn,sw,multi,maf,selective,htcn,crda} use adversarial learning to align features from both domains with a gradient reverse layer (GRL). Domain translation methods \cite{translation1,translation2,translation3} focus on translating source data into target-like styles to improve the performance of domain-adaptive object detection. Self-training methods \cite{umt,pt,at,cmt,ht,cat}, which have shown superior advantages in this field, use weak-strong augmentation and Mean Teacher (MT) \cite{mt} for teacher-student mutual learning. Adaptive Teacher (AT) \cite{at} additionally incorporates adversarial learning to bridge the domain gap. Contrastive Mean Teacher (CMT) \cite{cmt} leverages contrastive learning to optimize object-level features, while Class-Aware Teacher (CAT) \cite{cat} applies instance-level mixup to mitigate bias toward the majority class.

\subsection{Pseudo-labeling}
Most cutting-edge DAOD approaches \cite{umt,at,cmt,cat} use vanilla pseudo-labeling, where confident predictions are selected with a static, unified threshold. However, due to the imbalance in the labeled source data, the learned model typically over-represents the more frequently observed categories. Selecting pseudo-labels based on confidence can exacerbate the model’s existing biases. To address this, a dynamic and class-specific threshold has been proposed to mitigate bias in the pseudo-labels \cite{threshold1,threshold2,threshold3}. While this method helps balance the distribution between majority and minority pseudo-labels, it often introduces additional false positives. Additionally, some methods focus on selecting pseudo-labels based on the localization quality \cite{pt,ht}. These methods typically train a separate branch for regression uncertainty on the source domain, which cannot guarantee its effectiveness on the target domain. An important perspective that has not been fully researched is a learning-free approach to produce uncertainty estimates beyond those provided by neural networks. In this paper, we propose using adversarial attacks to detect and regularize pseudo-labels, offering a novel solution to this challenge.

\subsection{Adversarial Attack}
Adversarial attacks introduce imperceptible perturbations to input data, causing the model to make different predictions. The Fast Gradient Sign Method (FGSM), proposed in \cite{fgsm}, leverages the gradients of a neural network to design adversarial examples. Building upon this, Projected Gradient Descent (PGD) employs multi-step perturbations for a more powerful attack \cite{pgd}. Following these approaches, Alarab et al. estimate the classification uncertainty by applying adversarial attacks \cite{adv_for_uncertainty1, adv_for_uncertainty2}. Unlike earlier studies on uncertainty estimation, which apply perturbations to the model's parameters as in Bayesian approaches \cite{bayesian1, bayesian2}, adversarial attacks have shown effectiveness in producing uncertainty through perturbations of the input data. In this paper, we extend this concept to the DAOD setting, using adversarial attacks for pseudo-label regularization.

\section{Preliminaries}
\subsection{Problem Definition}
DAOD aims at mitigating the impact of domain shifts between the source domain $\mathcal{D}_s=\{X_s,B_s,C_s\}$ and the target domain $\mathcal{D}_t=\{X_t\}$ on object detectors. Source images $X_s$ are labeled with corresponding bounding box annotations $B_s$ and class labels $C_s$, while target images $X_t$ are not annotated.
\subsection{Mean Teacher}
State-of-the-art methods \cite{pt,at,cmt,ht,cat} employ the MT framework \cite{mt} for DAOD. A source model is first pre-trained on labeled source data, serving as the initial model for two architecturally identical models: the teacher and the student model. The teacher model generates pseudo-labels $B_t$ and $C_t$ with confidence higher than a hard threshold $\tau$ on weakly augmented target samples, while the student model is trained on both the labeled source data $\{X_s,B_s,C_s\}$ and strongly augmented target data $\{X_t,B_t,C_t\}$. The consistency loss between the pseudo-labels generated by the teacher model and the predictions of the student model improves the generalization capability of the student model through gradient back-propagation. Concurrently, the teacher model is updated through the EMA of the weights of the student model, performed without any gradient involvement:

\begin{equation}
    \theta_{\text{teacher}} \leftarrow \alpha \theta_{\text{teacher}} + (1-\alpha)\theta_{\text{student}}
    \label{eq:ema}
\end{equation}

Taking Faster R-CNN \cite{fasterrcnn} as an example, the optimization objective of the student model on labeled source domain and pseudo-labeled target domain can be respectively written as follows:
\begin{equation}
    \begin{aligned}
      \mathcal{L}_{s} = \mathcal{L}^{\text{rpn}}(X_s,B_s,C_s)+\mathcal{L}^{\text{roi}}(X_s,B_s,C_s)
    \end{aligned}
    \label{eq:ls}
\end{equation}

and 
\begin{equation}
    \begin{aligned}
      \mathcal{L}_{t} = \mathcal{L}^{\text{rpn}}_{\text{cls}}(X_t,B_t,C_t)+\mathcal{L}^{\text{roi}}_{\text{cls}}(X_t,B_t,C_t)
    \end{aligned}
    \label{eq:lt}
\end{equation}
Following \cite{at, cat} we exclude regression losses in $\mathcal{L}_t$ and implement a discriminator to learn domain invariant features with adversarial learning.
\subsection{Fast Gradient Sign Method}
FGSM stands as a widely employed adversarial attack method in image classification. Given an input image, FGSM uses the gradients of the loss concerning the input image to generate a new image that maximizes the classification loss. This generated adversarial example can be expressed as in Equation~\ref{fgsm}.
\begin{equation}
    x^{\text{adv}}=x+\beta\cdot \text{sgn}(\nabla_x \mathcal{L}_{cls}(\theta,x,y))
    \label{fgsm}
\end{equation}
where the parameter $\beta$ governs the magnitude of the adversarial perturbation, and $\text{sgn}(\cdot)$ refers to the sign function. Furthermore, $\mathcal{L}_{cls}(\theta,x,y)$ indicates the loss between the prediction and the ground truth $y$. 

\section{Proposed Method}
In this section, we introduce the proposed Adversarial Attacked Teacher (AAT) framework illustrated in Figure~\ref{fig2}. We first explain the process of conducting adversarial attacks on the teacher model, then provide a detailed discussion of the AAT design, including Adaptive Pseudo-label Regularization (APR) and Robust Minority Oversampling (RMO).
\subsection{Adversarial Attack on the Teacher Model}
We employ adversarial attacks on the teacher model to estimate the uncertainty of the pseudo-labels. The teacher model is encouraged to reassess pseudo-labels under the effect of a human imperceptible perturbation and to adjust its predictions if uncertain about the classification result. This is similar to how humans reevaluate their decisions when confronted with challenging or uncertain optical illusions, thereby improving the robustness and accuracy of the model's predictions. 

Instead of using the ground truth, which is unavailable for target data, we generate adversarial examples based on vanilla pseudo-labels. Here we only attack the classification branch of ROI since it generates the confidence score which is essential for pseudo-label selection. For each proposal that is matched as an object, the objective is to subtly modify the data so that the cross-entropy loss between its prediction $p_i$ and its vanilla pseudo-label $c_i$ is maximized. Equation~\ref{fgsm} is then adapted as:

\begin{equation}
    \begin{aligned}
      x^{\text{adv}}=x+\beta\cdot \text{sgn}(\nabla_x(\sum_{c_i\neq c_{\text{back}}}\mathcal{L}_{\text{cls}}(p_i,c_i)))
    \end{aligned}
    \label{eq:attack}
\end{equation}

\begin{figure}[t]
\centering
\includegraphics[width=\columnwidth]{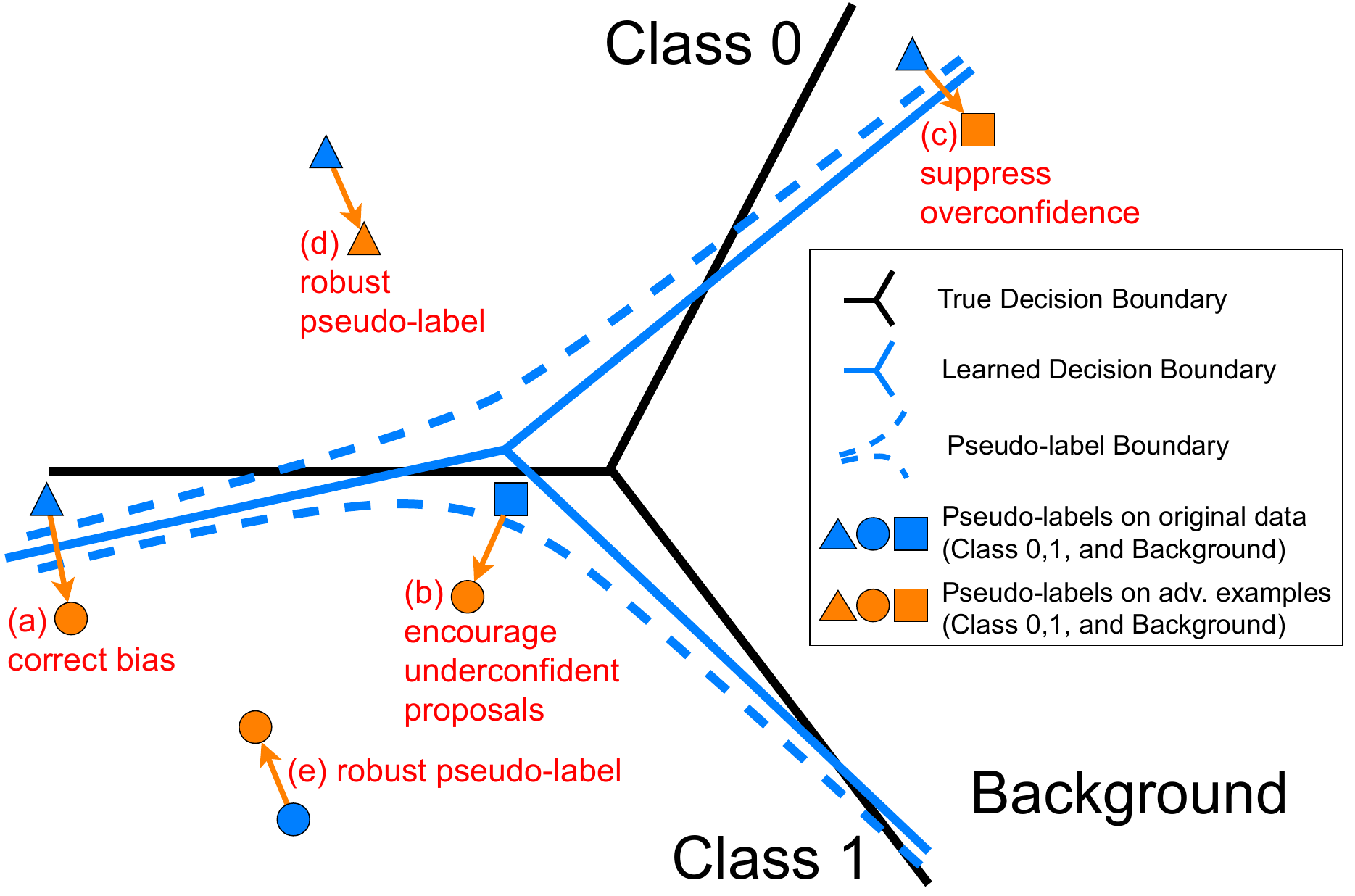} 
\caption{Illustrative examples of pseudo-labels on the \textit{original data} (blue) and \textit{the corresponding adversarial examples} (orange).}
\label{fig3}
\end{figure}

In Figure~\ref{fig3}, we illustrate various scenarios involving pseudo-labels under adversarial attacks. Within this context, Class 0 and Class 1 represent two distinct categories, with Class 0 being the more dominant class. Our experiments reveal that the model tends to overfit the minority class, Class 1, leading to a biased decision boundary that favors Class 0. In Figure~\ref{fig3}(a), we observe a sample from Class 1 that is incorrectly classified as Class 0 because of the skewed decision boundary. The teacher model, however, corrects this biased prediction when confronted with an adversarial example. In case (b), the sample is correctly classified but is excluded from pseudo-labels due to its low confidence score. Persistently omitting such underconfident minority class proposals from pseudo-labels exacerbates the skew in the decision boundary. Adversarial attacks mitigate this by prompting the teacher model to assign higher confidence scores to underconfident samples. Figure~\ref{fig3}(c) shows a situation where part of an object or the background receives an excessively high confidence score, potentially leading to false positives. Adversarial attacks help reduce this overconfidence. Conversely, in cases (d) and (e), where samples are distant from the decision boundary and resilient to adversarial attacks, the pseudo-labels are considered highly reliable.


\subsection{Adaptive Pseudo-label Regularization}
Pseudo-labels exhibit varying levels of certainty. Those with lower certainty are more prone to be incorrect, making it unreasonable to treat all pseudo-labels uniformly. To address this, we generate two sets of pseudo-labels with partial overlap: vanilla pseudo-labels generated by the standard pseudo-labeling process and adversarial pseudo-labels, which will be explained later. APR entails training the model to fit both pseudo-labels, thereby reinforcing the influence of the more reliable pseudo-labels while diminishing the impact of the less certain ones.

\begin{algorithm}[ht]
\caption{Adversarial Pseudo-label Generation}
\label{alg:algorithm}
\textbf{Input}: Vanilla pseudo-labels $\{B_t,C_t\}$ generated on original target data,  pseudo-labels $\{B_t',C_t'\}$ generated on adversarial examples and a confusion matrix $M$ representing class bias\\
\textbf{Output}: Adversarial pseudo-labels $\{B_t^{\text{adv}},C_t^{\text{adv}}\}$\\
\begin{algorithmic}[1] 
\STATE Let $\{B_t^{\text{adv}},C_t^{\text{adv}}\}=\emptyset$.
\FOR {$\{b_i',c_i'\}$ in $\{B_t',C_t'\}$}
\STATE find the best match $b_j$ for $b_i'$ in $B_t$.
\IF {$\text{IOU}(b_i',b_j) > 0.5$}
\IF {$M[c_i',c_j] \geq M[c_j,c_i']$}
\STATE add $\{b_i',c_i'\}$ to $\{B_t^{\text{adv}},C_t^{\text{adv}}\}$
\ENDIF
\ELSIF{$M[c_i',c_i']<\text{avg}(\text{diag}(M))$}
\STATE add $\{b_i',c_i'\}$ to $\{B_t^{\text{adv}},C_t^{\text{adv}}\}$ 
\ENDIF
\ENDFOR
\STATE \textbf{return} $\{B_t^{\text{adv}},C_t^{\text{adv}}\}$
\end{algorithmic}
\end{algorithm}
Based on the pseudo-labels generated on original data and adversarial examples, we construct adversarial pseudo-labels as in Algorithm~\ref{alg:algorithm}. Similar to in \cite{cat}, we maintain a confusion matrix $M$ in an exponential average manner on the source domain approximating the class dominance. $M[i,j]$ indicates the average probability of an instance of category $i$ being classified as category $j$. For each $(i, j)$ pair, we determine $i$ as a less dominant category than $j$ when $M[i,j] > M[j,i]$. Meanwhile, classes with $M[i,i]$ above and below the mean value $\text{avg}(\text{diag}(M))$ are recognized as majority and minority classes, respectively. Vanilla pseudo-labels robust against adversarial attacks are retained, while those reclassified as a less dominant category are corrected. Newly identified pseudo-labels for minority classes in adversarial examples are also included. The bottom row in Figure~\ref{fig1} illustrates the adversarial pseudo-labels corresponding to these samples.

The primary advantage of APR is its ability to generate appropriate soft labels for proposals where the teacher model exhibits uncertainty. For instance, if a proposal is classified as a car in the $\{B_t,C_t\}$ but as a truck in $\{B_t^{\text{adv}},C_t^{\text{adv}}\}$, the student model learns to predict a soft label that balances between car and truck, rather than committing to a potentially incorrect classification of car.

The overall objective is summarized in Equation~\ref{eq:l}:
\begin{equation}
    \begin{aligned}
      \mathcal{L} = \mathcal{L}_s+\lambda_t(\mathcal{L}_t + \mathcal{L}_t^{\text{adv}}) + \lambda_{\text{dis}}\mathcal{L}_{\text{dis}}
    \end{aligned}
    \label{eq:l}
\end{equation}
with
\begin{equation}
    \begin{aligned}
      \mathcal{L}_{t}^{\text{adv}} = \mathcal{L}^{\text{rpn}}_{\text{cls}}(X_t,B_t^{\text{adv}},C_t^{\text{adv}})+\mathcal{L}^{\text{roi}}_{\text{cls}}(X_t,B_t^{\text{adv}},C_t^{\text{adv}})
    \end{aligned}
    \label{eq:lt_adv}
\end{equation}

\subsection{Robust Minority Oversampling}
APR provides a reasonable soft label for the student model, but dataset imbalance remains a challenge. Even in supervised settings, models often exhibit bias towards majority classes. To address this, we propose oversampling robust minority objects to achieve a more balanced representation and enhance model performance.

As illustrated in Algorithm~\ref{alg:algorithm}, robust vanilla pseudo-labels are included in adversarial pseudo-labels if they withstand adversarial attacks. We crop these robust patches of minority classes and store them in a Crop Bank, which operates on a first-in-first-out basis. These crops are oversampled, augmented, and used to cover parts of the cutout area during strong augmentations on target data, with corresponding labels added. This approach helps mitigate class imbalance bias and reduces the risk of introducing false positives by only oversampling robust pseudo-labels.

\section{Experiments}
\subsection{Datasets}
To validate the effectiveness of our approach, we conduct experiments on multiple benchmarks following prior works \cite{at,cmt,cat}, including 1) adaptation from normal weather (Cityscapes) to foggy weather (Foggy Cityscapes), 2) adaptation from real-world (PASCAL VOC) to artistic (Clipart1k). The public datasets used in our experiments are as follows:
\paragraph{Cityscapes.} Cityscapes \cite{cityscapes} contains 2,975 training images and 500 validation images, all captured in normal weather conditions. Each image is annotated with pixel-level labels. We convert the instance segmentation labels into bounding box annotations for our experiments.
\paragraph{Foggy Cityscapes.} Foggy Cityscapes \cite{foggy_cityscapes} is generated by adding synthesized fog on images in the Cityscapes. Each image is rendered with three levels of fog (0.005, 0.01, 0.02), representing the visibility ranges of 600, 300, and 150 meters. We conduct our experiments on the 0.02 split with the most severe fog.
\paragraph{PASCAL VOC.} PASCAL VOC \cite{voc} is a real-world object detection dataset comprising images of objects across 20 categories. Following \cite{sw,at}, we use a combination of VOC 2007 and 2012 as labeled source data, resulting in a total of 16,551 images.
\paragraph{Clipart1k.} Clipart1k \cite{clipart} shares the same 20 categories but demonstrates a huge real-world to artistic domain shift with PASCAL VOC. In line with \cite{at,cmt,cat}, the 1,000 clipart images of Clipart1k are split into training and test sets, comprising 500 images each.

\subsection{Implementation Details}
Following prior studies, our Adversarial Attacked Teacher (AAT) relies on Faster R-CNN \cite{fasterrcnn} with VGG-16 \cite{vgg} and ResNet-101 \cite{resnet} as the foundational detection model, implemented using the Detectron2 framework \cite{detectron2}. And these backbones are pre-trained on ImageNet \cite{imagenet}. A confidence threshold of $\delta=0.8$ is set for all experiments, and optimization is performed using Stochastic Gradient Descent (SGD). Data augmentation includes random horizontal flips for weak augmentation, and strong augmentations involve random color jittering, grayscaling, Gaussian blurring, and Cutout. Similar to CMT, we observe some objects completely erased by Cutout. We remove the objects from the pseudo-labels if 80\% of the proposal area is erased. The weight smoothing factor for the EMA ($\beta$ in \ref{eq:ema}) is set to 0.9996.  We report the average precision (AP) with a threshold of 0.5 for each class as well as the mean AP (mAP) over all classes for object detection following existing works for all of the experimental settings.

\subsection{Adverse Weather Adaptation}
Fog, a meteorological occurrence, occurs when water droplets or ice crystals gather in the air close to the ground, resulting in diminished visibility. Its impact on visibility, such as scattering and absorption of light, causes distant objects to appear blurry or obscured. In this experiment, we evaluate AAT on the commonly used benchmark Cityscapes $\rightarrow$ Foggy Cityscapes, where the object detector needs to overcome the domain shift from normal to foggy weather.

\begin{table}[ht]
\fontsize{9pt}{9pt}\selectfont
\setlength{\tabcolsep}{1mm}
\begin{tabular}{@{}c|cccccccc|c@{}}
\toprule
Method    & prsn        & rider         & car           & truck         & bus           & train         & mtr        & bike       & mAP           \\ \midrule
Source    & 22.4          & 26.6          & 28.5          & 9.0           & 16.0          & 4.3           & 15.2          & 25.3          & 18.4          \\
Oracle    & 39.5          & 47.3          & 59.1          & 33.1          & 47.3          & 42.9          & 38.1          & 40.8          & 43.5          \\ \midrule
DA-Faster & 25.0          & 31.0          & 40.5          & 22.1          & 35.3          & 20.2          & 20.2          & 27.1          & 27.6          \\
SW        & 29.9          & 42.3          & 43.5          & 24.5          & 36.2          & 32.6          & 30.0          & 35.3          & 34.3          \\
DM        & 30.8          & 40.5          & 44.3          & 27.2          & 38.4          & 34.5          & 28.4          & 32.2          & 34.6          \\
HTCN      & 33.2          & 47.5          & 47.9          & 31.6          & 47.4          & 40.9          & 32.3          & 37.1          & 39.8          \\
UMT       & 33.0          & 46.7          & 48.6          & 34.1          & 56.5          & 46.8          & 30.4          & 37.4          & 41.7          \\
PT        & 40.2          & 48.8          & 59.7          & 30.7          & 51.8          & 30.6          & 35.4          & 44.5          & 42.7          \\
TDD       & 39.6          & 47.5          & 55.7          & 33.8          & 47.6          & 42.1          & 37.0          & 41.4          & 43.1          \\
AT\dag    & 45.3          & 55.7          & 63.6          & 36.8          & 64.9          & 34.9          & 42.1          & 51.3          & 49.3          \\
CMT       & \underline{45.9}    & 55.7          & 63.7          & \underline{39.6}    & \textbf{66.0} & 38.8          & 41.4          & 51.2          & 50.3          \\
HT        & \textbf{52.1} & 55.8          & \textbf{67.5} & 32.7          & 55.9          & 49.1          & 40.1          & 50.3          & 50.4          \\
CAT       & 44.6          & \underline{57.1}    & 63.7          & \textbf{40.8} & \textbf{66.0} & \underline{49.7}    & \underline{44.9}    & \underline{53.0}    & \underline{52.5}    \\ \midrule
AAT       & 44.8          & \textbf{57.5} & \underline{64.0}    & 37.4          & 63.6          & \textbf{57.0} & \textbf{45.2} & \textbf{54.8} & \textbf{53.0} \\ \bottomrule
\end{tabular}%
\caption{Results of \textit{Cityscapes $\rightarrow$ Foggy Cityscapes}. \dag AT performance is reproduced with publicly available code.}
\label{tab:c2f}
\end{table}

The results are summarized in Table~\ref{tab:c2f}. We present the evaluation results, comparing them with the performance of source (fully supervised on the source domain) and oracle models (fully supervised on the target domain). Similar to many other Mean Teacher-based approaches (AT, CMT, HT, CAT), our method surpasses the performance of the oracle models. Furthermore, our AAT outperforms all state-of-the-art approaches and reaches 53.0 mAP. Compared to our baseline, AAT shows a significant improvement in the detection performance of minority categories, including train (+22.1 AP), motorcycle (+3.1 AP), and rider (+1.8 AP). This validates the effectiveness of our approach in correcting bias and enhancing the model's ability to accurately detect underrepresented objects in challenging domains.

\subsection{Artistic Adaptation}
Real-world objects and clipart images differ significantly in visual complexity and style, challenging domain adaptation. Adapting from real-world to clipart images is crucial for applications in creative industries, where content often transitions between realistic and stylized forms. We evaluate AAT on the commonly used benchmark PASCAL VOC $\rightarrow$ Clipart1k, where the object detector needs to overcome the domain shift from real-world to artistic. The results are shown in Table~\ref{tab:p2c}. The proposed AAT achieves a mAP of 52.6, surpassing the previous SOTA by 6.7\%. Notably, AAT exhibits substantial improvement in minority classes compared to the baseline, achieving a +12.9 AP increase in the motorbike category, which is represented by only 7 instances in the Clipart1k training set, and a +15.7 AP increase in the bus category, which contains only 13 instances.
\begin{table*}[ht]
\fontsize{9pt}{9pt}\selectfont
\setlength{\tabcolsep}{1mm}
\begin{tabular}{@{}c|cccccccccccccccccccc|c@{}}
\toprule
Method    & aero          & bike          & bird          & boat          & bottle        & bus           & car           & cat           & chair         & cow           & table         & dog           & horse         & mtr           & prsn          & plant         & shp           & sofa          & train         & tv            & mAP           \\ \midrule
Source    & 23.0          & 39.6          & 20.1          & 23.6          & 25.7          & 42.6          & 25.2          & 0.9           & 41.2          & 25.6          & 23.7          & 11.2          & 28.2          & 49.5          & 45.2          & 46.9          & 9.1           & 22.3          & 38.9          & 31.5          & 28.8          \\
Oracle    & 33.3          & 47.6          & 43.1          & 38.0          & 24.5          & 82.0          & 57.4          & 22.9          & 48.4          & 49.2          & 37.9          & 46.4          & 41.1          & 54.0          & 73.7          & 39.5          & 36.7          & 19.1          & 53.2          & 52.9          & 45.0          \\ \midrule
SW        & 26.2          & 48.5          & 32.6          & 33.7          & 38.5          & 54.3          & 37.1          & \textbf{18.6} & 34.8          & 58.3          & 17.0          & 12.5          & 33.8          & 65.5          & 61.6          & 52.0          & 9.3           & 34.9          & \underline{54.1}    & 49.1          & 38.1          \\
DM        & 25.8          & 63.2          & 24.5          & \underline{42.4}    & 47.9          & 43.1          & 37.5          & 9.1           & 47.0          & 46.7          & 26.8          & 24.9          & 48.1          & 78.7          & 63.0          & 45.0          & 21.3          & 36.1          & 52.3          & \underline{53.4}          & 41.8          \\
HTCN      & 33.6          & 58.9          & 34.0          & 23.4          & 45.6          & 57.0          & 39.8          & 12.0          & 39.7          & 51.3          & 21.1          & 20.1          & 39.1          & 72.8          & 63.0          & 43.1          & 19.3          & 30.1          & 50.2          & 51.8    & 40.3          \\
I3Net     & 30.0          & \textbf{67.0} & 32.5          & 21.8          & 29.2          & \underline{62.5}          & 41.3          & 11.6          & 37.1          & 39.4          & 27.4          & 19.3          & 25.0          & 67.4          & 55.2          & 42.9          & 19.5          & 36.2          & 50.7          & 39.3          & 37.8          \\
UMT       & 39.6          & 59.1          & 32.4          & 35.0          & 45.1          & 61.9          & 48.4          & 7.5           & 46.0          & \underline{67.6}    & 21.4          & \underline{29.5}    & 48.2          & 75.9          & 70.5          & 56.7          & 25.9          & 28.9          & 39.4          & 43.6          & 44.1          \\
AT        & 33.8          & 60.9          & 38.6          & \textbf{49.4} & 52.4          & 53.9          & \underline{56.7}    & 7.5           & 52.8          & 63.5          & 34.0          & 25.0          & 62.2          & 72.1          & 77.2          & \underline{57.7}    & \underline{27.2}    & \textbf{52.0} & \textbf{55.7} & \textbf{54.1} & \underline{49.3}    \\
CMT       & 39.8          & 56.3          & 38.7          & 39.7          & \underline{60.4}    & 35.0          & 56.0          & 7.1           & \textbf{60.1} & 60.4          & 35.8          & 28.1          & \textbf{67.8} & 84.5          & \textbf{80.1} & 55.5          & 20.3          & 32.8          & 42.3          & 38.2          & 47.0          \\
CAT       & \underline{40.5}    & 64.1          & \underline{38.8}    & 41.0          & \textbf{60.7} & 55.5          & 55.6          & 14.3          & 54.7          & 59.6          & \textbf{46.2} & 20.3          & 58.7          & \textbf{92.9} & 62.6          & 57.5          & 22.4          & \underline{40.9}    & 49.5          & 46.0          & 49.1          \\ \midrule
AAT & \textbf{46.5} & \underline{65.0}    & \textbf{40.1} & 34.6          & 60.0          & \textbf{69.6} & \textbf{58.5} & \underline{14.9}    & \underline{59.8}    & \textbf{70.6} & \underline{38.9}    & \textbf{40.1} & \underline{65.2}    & \underline{85.0}    & \underline{78.2}    & \textbf{58.8} & \textbf{41.8} & 38.0          & 41.0          & 45.7          & \textbf{52.6} \\ \bottomrule
\end{tabular}%
\caption{Results of \textit{PASCAL VOC $\rightarrow$ Clipart1k}.}
\label{tab:p2c}
\end{table*}
\subsection{Ablation Studies}
To assess the impact of our contributions, we performed an ablation study. All experiments in this study were conducted on the PASCAL VOC $\rightarrow$ Clipart1k benchmark, using the ResNet-101 backbone.
\paragraph{Adaptive Pseudo-label Regularization.} The sole hyperparameter introduced by APR is the step length $\beta$ of the adversarial attacks, as defined in Equation~\ref{eq:attack}. This parameter controls the intensity of the adversarial perturbation, thereby influencing the extent to which the teacher model's vanilla pseudo-labels are regularized.

\begin{figure}[ht]
\centering
\includegraphics[width=\columnwidth]{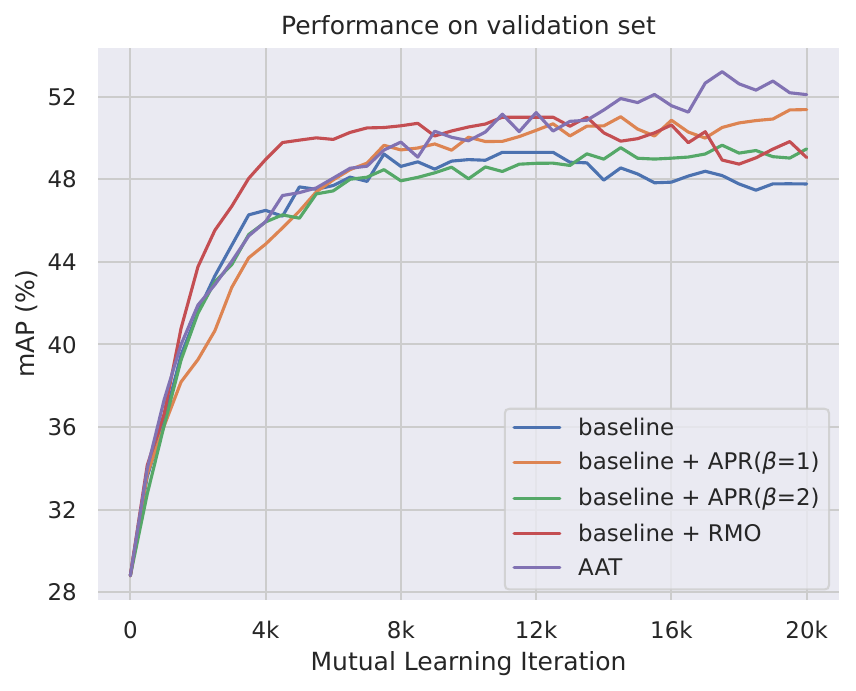} 
\caption{Learning curves on Clipart1k dataset.}
\label{fig4}
\end{figure}

Experimental results validating the effectiveness of this approach are illustrated in Figure~\ref{fig4}, where different values of $\beta$ are tested to facilitate the baseline (AT). Previous studies have observed that domain adaptive approaches typically show initial performance improvements, which can later decline as the teacher model becomes increasingly biased and overconfident. This issue is also evident in AT. For example, the baseline model's performance (blue line) initially improves to 49.3 mAP but then drops, indicating insufficient adaptation. This performance decline highlights the necessity of adaptive pseudo-label regularization, which reinforces reliable pseudo-labels and mitigates the impact of uncertain ones. With an optimal regularization level ($\beta=1$), performance steadily improves to 51.4 mAP, indicating that the model learns a less biased representation and maintains performance gains. Conversely, if $\beta$ is set too high ($\beta=2$), the model becomes excessively cautious, accepting only highly certain pseudo-labels. This restricts learning from potentially useful but less confident data, resulting in a lower performance of 49.6 mAP.

\begin{table}[ht]
\centering
\begin{tabular}{@{}c|cc|cc}
\toprule
Methods                     & APR                  & RMO                             & mAP  & Gain w.r.t. AT \\ \midrule
Baseline (AT)               &                      &                                 & 49.3 & -              \\ \midrule
\multirow{3}{*}{AAT (Ours)} & \checkmark           &                                 & 51.4 & 2.1            \\
                            &                      & \checkmark                      & 51.0 & 1.7            \\
                            & \checkmark           & \checkmark                      & 52.6 & 3.3            \\ \bottomrule
\end{tabular}%
\caption{Ablation Studies. RMO without APR refers to oversampling all minority categories in the vanilla pseudo-labels without checking their reliability.}
\label{tab:ablation}
\end{table}

\paragraph{Robust Minority Oversampling.} Datasets often suffer from significant category imbalances, with certain classes being vastly overrepresented compared to others. To combat this issue, oversampling is a commonly used technique, where instances of the minority class are replicated to ensure they appear more frequently during training. However, in an unsupervised domain adaptation setting, the effectiveness of oversampling in the target domain heavily depends on the pseudo-labels' quality. If the pseudo-labels are inaccurate or biased, oversampling can amplify these errors, leading to poor model performance. Therefore, we propose oversampling only those robust pseudo-labels verified by APR. Results in Table~\ref{tab:ablation} show that combining APR and RMO yields a 3.3 mAP improvement over AT. In contrast, oversampling all minority pseudo-labels without assessing their reliability with APR results in only a 1.7 mAP improvement. While this approach initially accelerates adaptation as shown in Figure~\ref{fig4}, the performance eventually declines, resembling that of the baseline. On the one hand, this validates the effectiveness of minority oversampling in addressing class imbalance; on the other hand, it underscores the importance of APR in ensuring that the pseudo-labels being oversampled are of high quality, thereby maximizing the overall performance gains.
\subsection{Qualitative Results}
\begin{figure}[t]
\centering
\includegraphics[width=0.95\columnwidth]{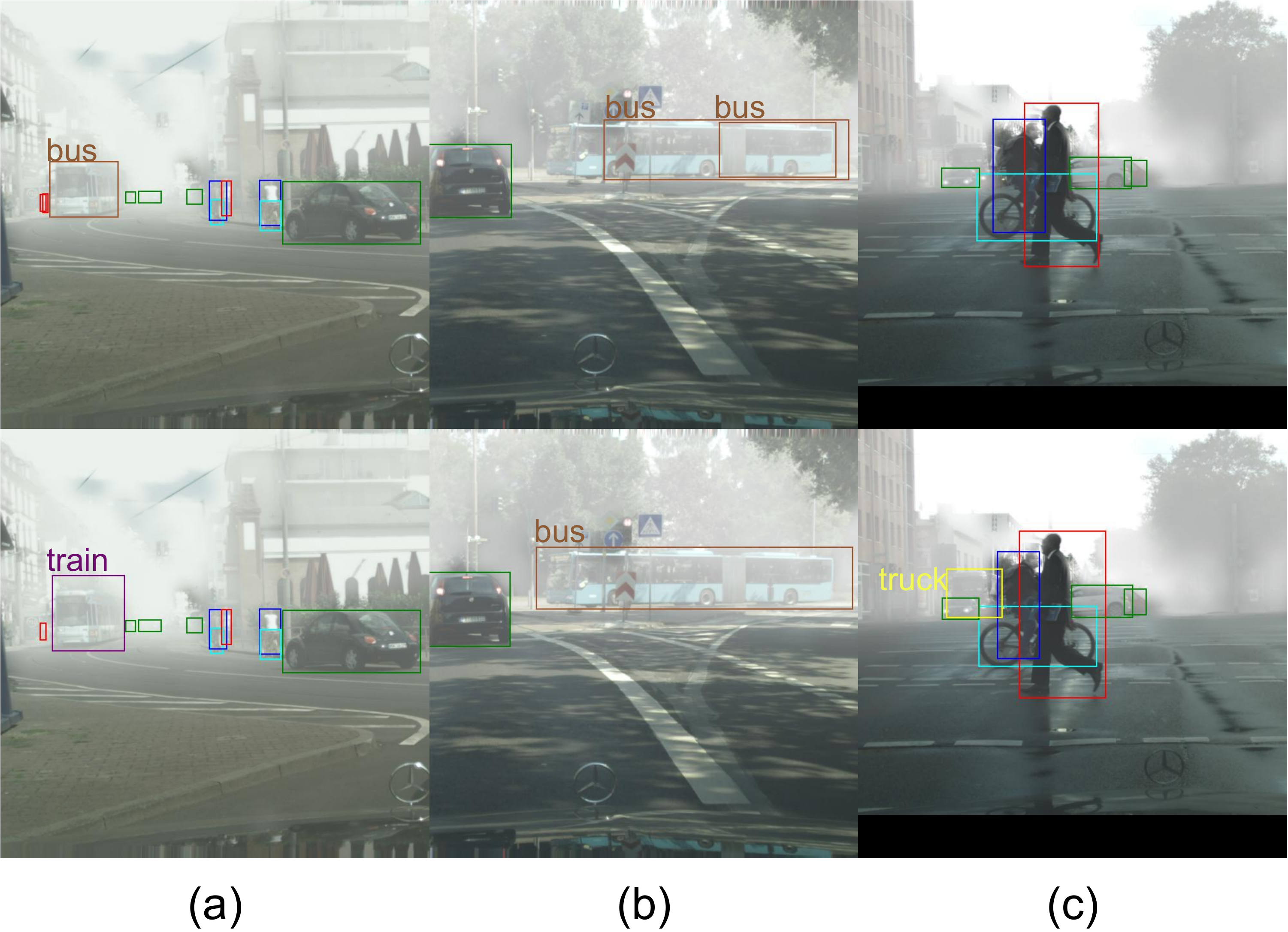} 
\caption{Qualitative results on Foggy Cityscapes. We compare AT (top) and AAT (bottom). AAT fixes (a) bias towards major classes, (b) false positives, and (c) false negatives.}
\label{fig5}
\end{figure}

Equipped with APR and RMO, the proposed AAT framework demonstrates the capability to correct misclassifications and reduce both false positives and false negatives, thereby improving overall detection performance. We compare AT and AAT on the Cityscapes $\rightarrow$ Foggy Cityscapes benchmark and provide qualitative results in Figure~\ref{fig5}.

\section{Conclusions}
In this paper, we address the critical challenge of domain shifts in object detection by introducing the AAT framework. Our work demonstrates that existing domain adaptive methods often suffer from biased pseudo-labels that can introduce misclassification, overconfident false positives, and underconfident false negatives. To tackle these issues, we propose a novel approach that leverages adversarial attacks to estimate the uncertainty and generate adversarial pseudo-labels as complementary labels. By jointly training with both vanilla and adversarial pseudo-labels, the model is regularized to emphasize reliable labels and mitigate the impact of noisy or uncertain predictions. This approach enhances the model’s ability to learn from high-certainty pseudo-labels while mitigating the risks associated with unreliable pseudo-labels. To further address class imbalances, we oversample robust minority objects verified by our APR, effectively balancing the dataset. Extensive experiments on multiple datasets validate the effectiveness of the AAT framework, with notable performance improvements. 
\bibliography{main}
\onecolumn
\section{Technical Appendix}



\subsection{Additional Visualization Results}
\begin{figure}[h]
    \centering
    \begin{subfigure}{0.21\textwidth}
        \centering
        \includegraphics[width=\textwidth]{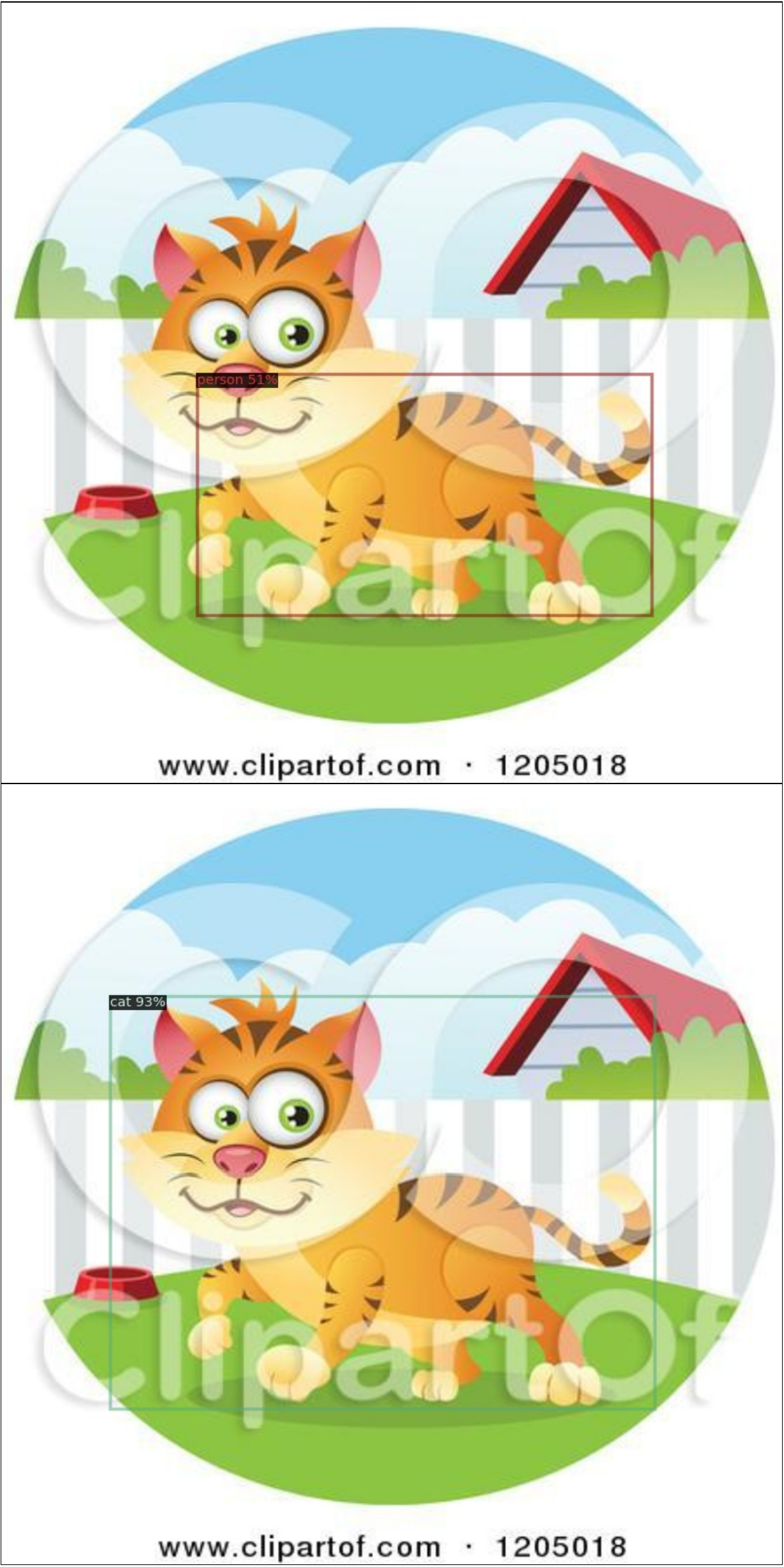}
        \caption{AAT fixes the misclassification of the cat.}
        \label{fig:subfig1}
    \end{subfigure}
    \hspace{0.1\textwidth}
    \begin{subfigure}{0.21\textwidth}
        \centering
        \includegraphics[width=\textwidth]{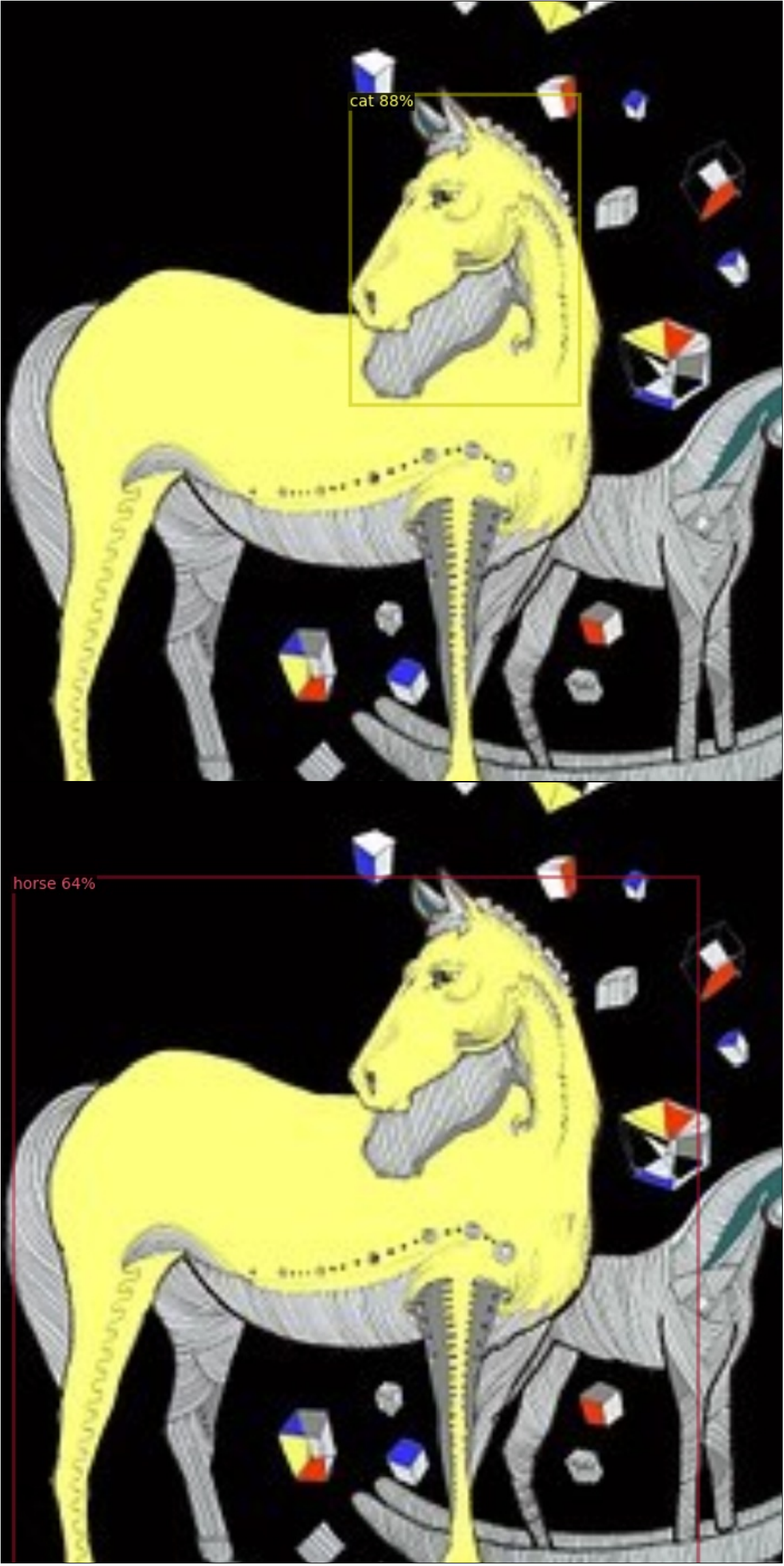}
        \caption{AAT fixes the misclassification of the horse.}
        \label{fig:subfig2}
    \end{subfigure}
    \hspace{0.1\textwidth}
    \begin{subfigure}{0.21\textwidth}
        \centering
        \includegraphics[width=\textwidth]{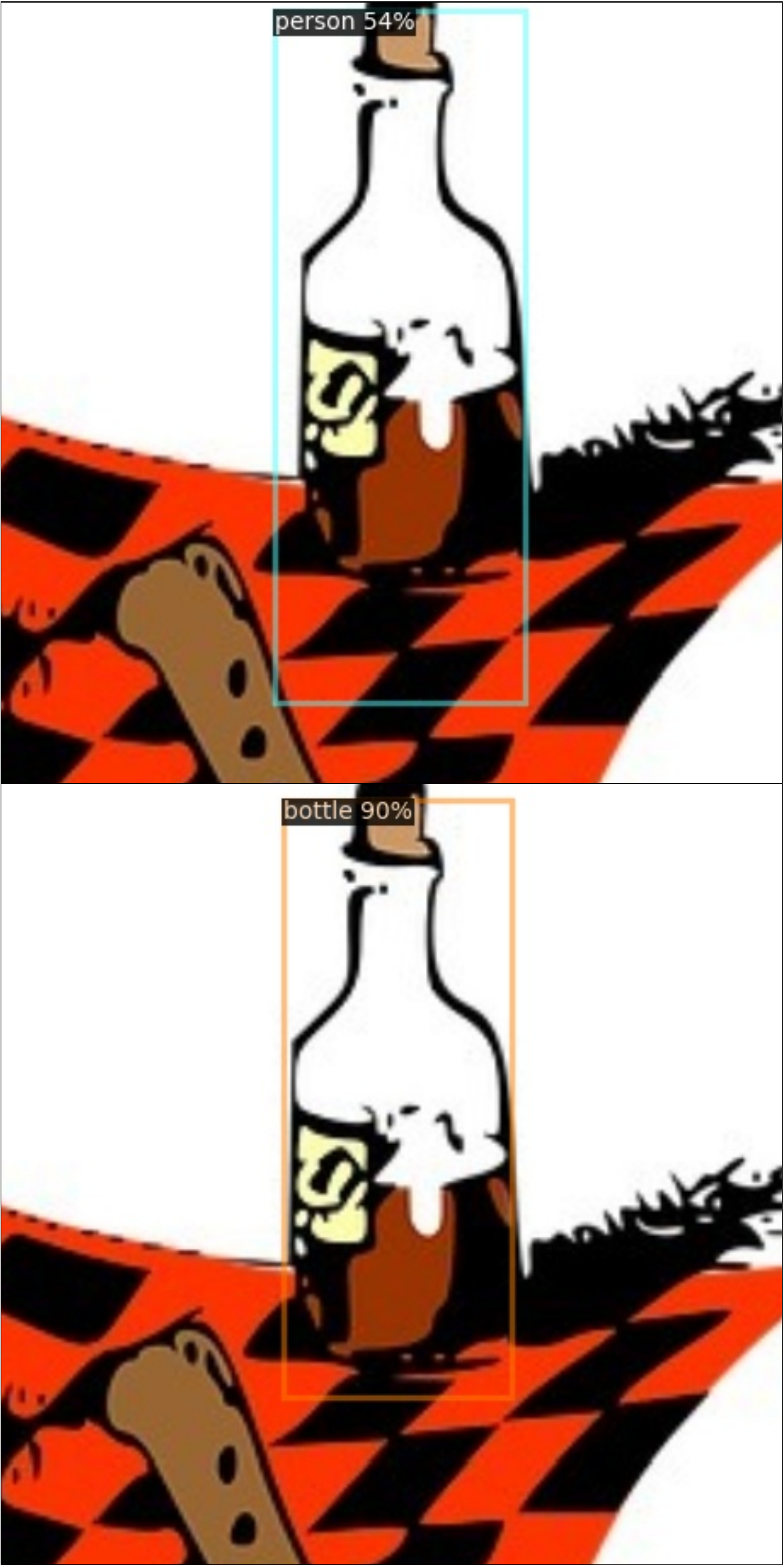}
        \caption{AAT fixes the misclassification of the bottle.}
        \label{fig:subfig3}
    \end{subfigure}

    \begin{subfigure}{0.21\textwidth}
        \centering
        \includegraphics[width=\textwidth]{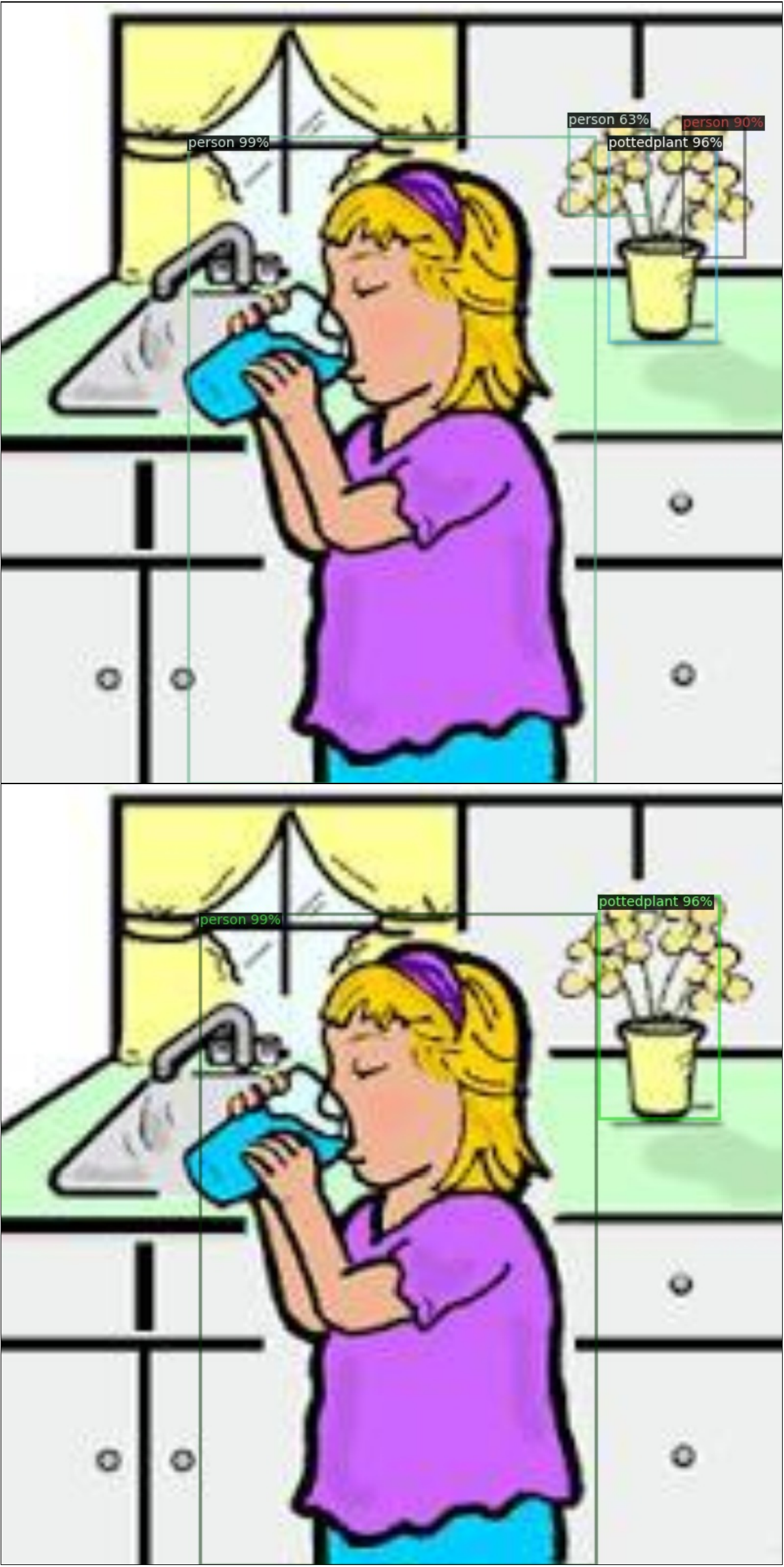}
        \caption{AAT avoids multiple false positives.}
        \label{fig:subfig4}
    \end{subfigure}
    \hspace{0.1\textwidth}
    \begin{subfigure}{0.21\textwidth}
        \centering
        \includegraphics[width=\textwidth]{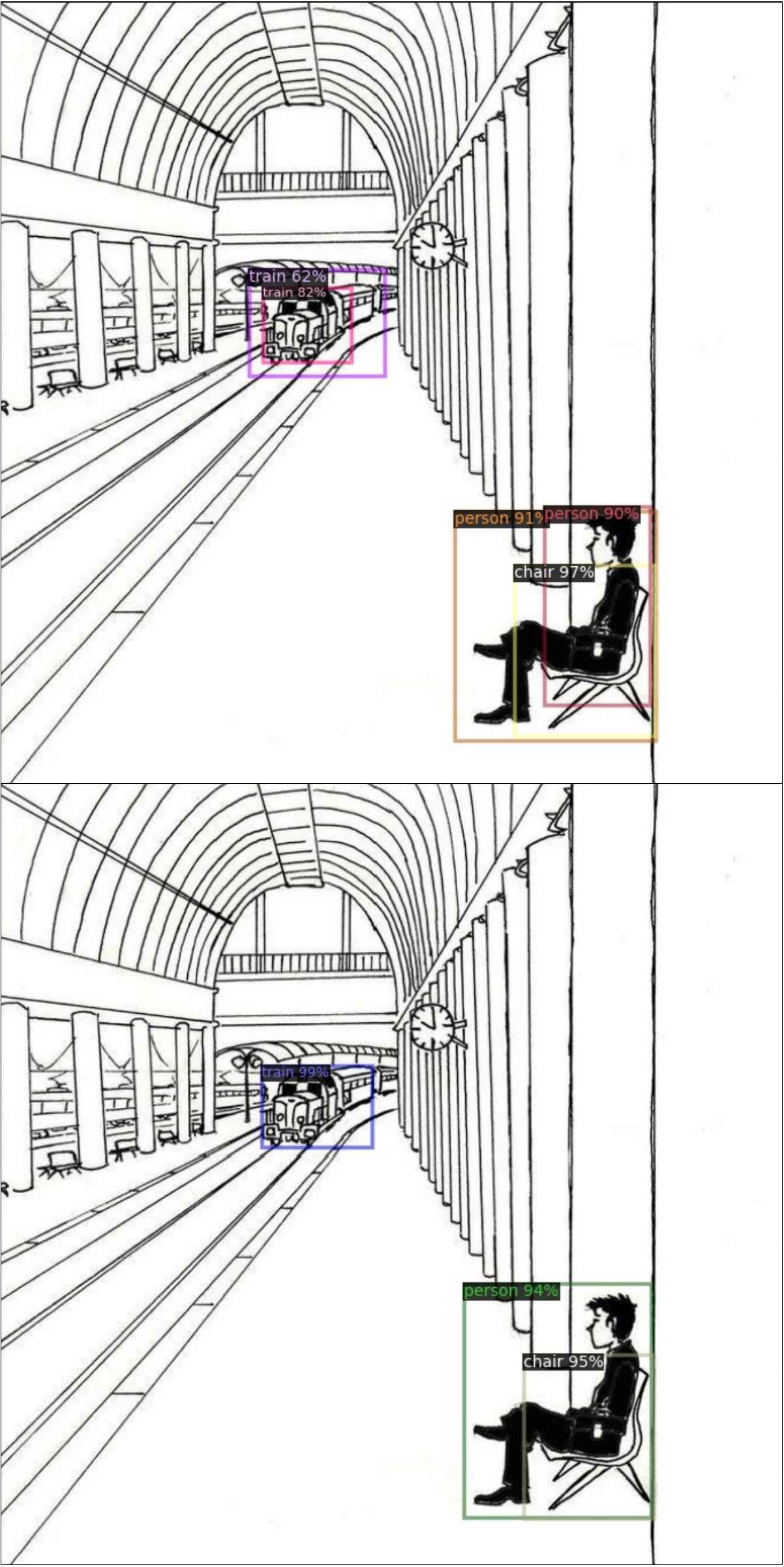}
        \caption{AAT avoids detecting multiple false positives.}
        \label{fig:subfig5}
    \end{subfigure}
    \hspace{0.1\textwidth}
    \begin{subfigure}{0.21\textwidth}
        \centering
        \includegraphics[width=\textwidth]{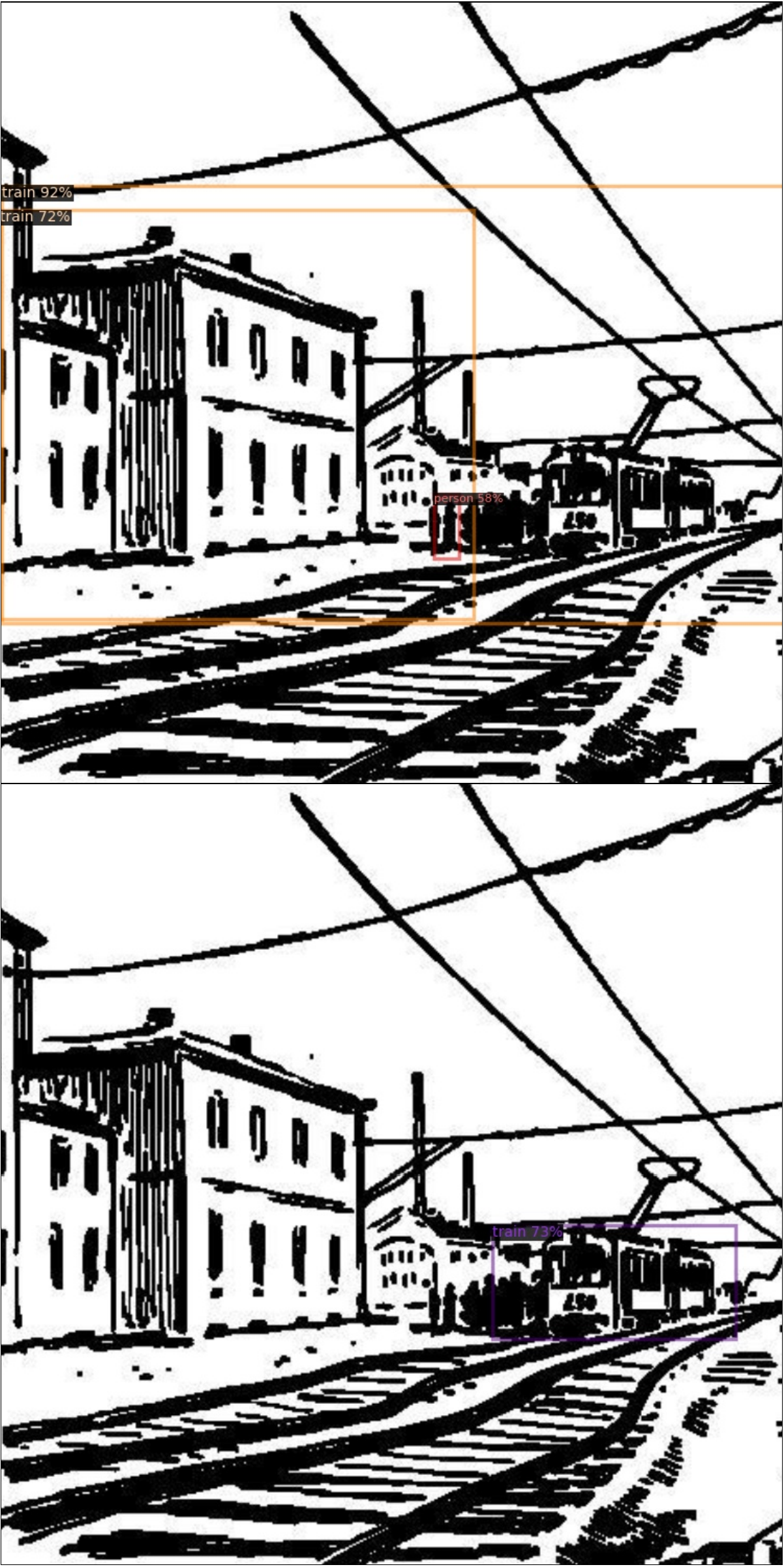}
        \caption{AAT detects the missing train.}
        \label{fig:subfig6}
    \end{subfigure}

    \caption{Additional qualitative results on Clipart1k. We compare AT (top) and AAT (bottom).}
    \label{fig6}
\end{figure}

We provide additional visualization results to compare the baseline Adaptative Teacher (AT) with our AAT on the Pascal VOC $\rightarrow$ Clipart1k benchmark in Figure~\ref{fig6}. Each pair of images shows results by AT (top) and AAT (bottom).

\subsection{Code}
The Code for the proposed Adversarial Attacked Teacher (AAT) and all corresponding experiments is available at https://anonymous.4open.science/r/AAT-2CA0.

\end{document}